\documentclass[sigconf]{acmart}

\usepackage{algorithm}
\usepackage{algorithmic}
\usepackage{multirow}
\usepackage{bm}
\usepackage{subfigure}
\usepackage{graphicx}

\copyrightyear{2022} 
\acmYear{2022} 
\setcopyright{acmlicensed}\acmConference[KDD '22]{Proceedings of the 28th ACM SIGKDD Conference on Knowledge Discovery and Data Mining}{August 14--18, 2022}{Washington, DC, USA}
\acmBooktitle{Proceedings of the 28th ACM SIGKDD Conference on Knowledge Discovery and Data Mining (KDD '22), August 14--18, 2022, Washington, DC, USA}
\acmPrice{15.00}
\acmDOI{10.1145/3534678.3539368}
\acmISBN{978-1-4503-9385-0/22/08}

\begin{document}

\title{Unified 2D and 3D Pre-Training of Molecular Representations}

\author{Jinhua Zhu}
\authornote{This work was done when Jinhua Zhu was an intern at Microsoft Research Asia.}
\email{teslazhu@mail.ustc.edu.cn}
\affiliation{
\institution{University of Science and Technology of China}
\city{Hefei}
\state{Anhui}
\country{China}
}
\author{Yingce Xia}
\authornote{Corresponding author.}
\email{yingce.xia@microsoft.com}
\affiliation{%
  \institution{Microsoft Research Asia}
  \city{Beijing}
  \country{China}}
  
\author{Lijun Wu}
\email{lijuwu@microsoft.com}
\affiliation{%
  \institution{Microsoft Research Asia}
  \city{Beijing}
  \country{China}}
  
\author{Shufang Xie}
\email{shufxi@microsoft.com}
\affiliation{%
  \institution{Microsoft Research Asia}
  \city{Beijing}
  \country{China}}
  
 \author{Tao Qin}
\email{taoqin@microsoft.com}
\affiliation{%
  \institution{Microsoft Research Asia}
  \city{Beijing}
  \country{China}}
  
   \author{Wengang Zhou}
\email{zhwg@ustc.edu.cn}
\affiliation{
\institution{University of Science and Technology of China}
\city{Hefei}
\state{Anhui}
\country{China}
}
\author{Houqiang Li}
\email{lihq@ustc.edu.cn}
\affiliation{
\institution{University of Science and Technology of China}
\city{Hefei}
\state{Anhui}
\country{China}
}

\author{Tie-Yan Liu}
\email{tyliu@microsoft.com}
\affiliation{%
\institution{Microsoft Research Asia}
\city{Beijing}
\country{China}}

\renewcommand{\shortauthors}{Jinhua Zhu et al.} 

\begin{abstract}
Molecular representation learning has attracted much attention recently. A molecule can be viewed as a 2D graph with nodes/atoms connected by edges/bonds, and can also be represented by a 3D conformation with 3-dimensional coordinates of all atoms. We note that most previous work handles 2D and 3D information separately, while jointly leveraging these two sources may foster a more informative representation.
In this work, we explore this appealing idea and propose a new representation learning method based on a unified 2D and 3D pre-training. 
Atom coordinates and interatomic distances are encoded and then fused with atomic representations through graph neural networks. The model is pre-trained on three tasks: reconstruction of masked atoms and coordinates, 3D conformation generation conditioned on 2D graph, and 2D graph generation conditioned on 3D conformation. We evaluate 
our method on $11$ downstream molecular property prediction tasks: $7$ with 2D information only and $4$ with both 2D and 3D information. Our method achieves state-of-the-art results on $10$ tasks, 
and the average improvement on 2D-only tasks 
is $8.3\%$. Our method also achieves significant improvement  on two 3D conformation generation tasks. 
\end{abstract}



\maketitle

\section{Introduction}
\label{sec:intro}
Deep learning techniques have attracted more and more attention recently in drug discovery~\citep{stokes2020deep,Jine2105070118}, bioinformatics \citep{Sharan2021} and cheminformatics~\cite{Tetko2020}. Obtaining effective molecular representations, usually high dimensional vectors that are friendly to neural network models, is a key prerequisite for deep learning based methods~\cite{devlin2018bert,liu2019roberta}.

Molecules can be expressed in different forms, including 2D molecular graphs and 3D molecular  conformations.  A 2D molecular graph describes the 2D topological structure of a molecule, where the nodes/atoms are connected by edges/bonds respectively. Many methods have been proposed to deal with molecular graphs such as  Graphormer~\citep{ying2021graphormer}, virtual node~\citep{pmlr-v70-gilmer17a}, etc. A 3D molecular conformation is represented by a set of coordinates for all the atoms of a molecule. People also propose various models based on 3D conformations, such as SchNet \citep{schnet2017}, DimeNet \cite{klicpera_dimenet_2020}, PAiNN \citep{PAINN_schutt21a}, etc. 

While the two types of representations are complementary, i.e., 2D graphs focus on topological connections of atoms and 3D conformations focus on spatial arranges of atoms, only limited works leverage them together. \citet{chen2021algebraic} use the multiscale weighted colored algebraic graphs (AG) to encode 3D conformation and obtain corresponding 3D representations. 
Besides, they use a bidirectional Transformer \cite{vaswani2017Transformer,devlin2018bert} to encode the molecular SMILES \cite{doi:10.1021/ci00057a005} (obtained by traversing a 2D molecular graph using depth-first-search) and obtain another representation. The two representations are fused together for downstream tasks. \citet{graphmvp} and \citet{stark20213d} are two recent works that use both 2D information and 3D information for molecule pre-training. Their common training objective is to maximize the mutual information between the 2D and 3D views of a molecule, where the 2D view and 3D view are encoded using two different modules.

Different from above methods~\cite{chen2021algebraic,graphmvp,DBLP:journals/corr/abs-2110-04126}, in this work, we propose a unified method that processes both 2D and 3D information of molecules in a single model, inspired by the recent trend and success of multi-modality modeling in deep learning research~\cite{pmlr-v139-ramesh21a,DBLP:journals/corr/KaiserGSVPJU17}. For examples, DALL-E \cite{pmlr-v139-ramesh21a} is a unified model that can encode images and texts together, and demonstrates its great power in text-to-image generation; \citet{DBLP:journals/corr/KaiserGSVPJU17} use one unified model for image, speech and text processing, and show that the tasks with less data benefit largely from the joint training.

Our model is  pre-trained on PCQM4Mv2~\citep{hu2021ogblsc}, where molecules are represented by both 2D graphs and corresponding 3D conformations. In our model, atomic coordinates and interatomic distances are encoded by a feed-forward network and then fused by a graph neural network. To effectively unify 2D and 3D information, we design several pre-training tasks: 

(1) Reconstruction of masked atoms and coordinates, which is to reconstruct randomly masked atoms and coordinates based on unmasked ones; 

(2) 3D conformation generation conditioned on 2D graph, which is to generate 3D conformation based on the 2D graph of a molecule; 

(3) 2D graph generation conditioned on 3D conformation, which is to generate 2D graph based on the 3D conformation of a molecule.

We use masked language modeling loss \citep{devlin2018bert,liu2019roberta} to reconstruct the masked atoms.
Besides, we adopt a permutation invariant loss function of symmetric substructures so that the training process is more effective. This is because the coordinates of atoms in symmetric substructures can be swapped, and conventional loss function cannot maintain permutation invariance. We also adopt the roto-translation invariance loss to 3D conformation generation conditioned on the 2D graph. Note that our model is compatible with molecules with 2D information only, where their coordinates can be randomly initialized.

We test our method on the following tasks: 

(1) Six molecular property prediction tasks from MoleculeNet~\cite{Wu2017moleculenet} and one molecular prediction tasks from OGB benchmark\footnote{\url{https://ogb.stanford.edu/docs/leader_graphprop/}} \cite{ogb-smallscale}.
The 3D conformations of these molecules are not accessible.

(2) Four toxicity prediction tasks from \cite{doi:10.1021/acs.jcim.7b00558,chen2021algebraic} where both 2D and 3D information are available.

(3) Two 3D molecular conformation tasks where the data is sampled from the Geometric Ensemble Of Molecules dataset~\cite{axelrod2020geom,shi2021learning}.


We achieve state-of-the-art results on $10$ out of the $11$ molecular property prediction tasks. Specifically, on the 2D-only tasks, we improve the previous best method by $8.3\%$ on average. For the toxicity prediction with 3D conformations, our method outperforms the previous deep learning based methods and the manually designed 3D molecular fingerprints. For the two conformation generation tasks on GEOM-QM9 and GEOM-Drugs, in terms of the mean matching score, we improve the previous best results by $7.7\%$ and $3.6\%$.

Our contributions can be summarized as follows:

(1) We propose a new method that jointly encodes both 2D and 3D information of molecules in a unified model. The learnt model can be used for both molecular prediction and conformation generation tasks.

(2) We propose several new training objective functions to fully utilize the data, which shows new directions of molecule pre-training.

(3) We achieve state-of-the-art results on 10 molecular prediction tasks and 2 conformation generation tasks. 

(4) We release our code and pre-trained models at the Github repository \url{https://github.com/teslacool/UnifiedMolPretrain} for reproducibility.

\section{Related Work}

\subsection{Molecular Pre-training with 2D Information}

Inspired by its success in natural language processing and computer vision, pre-training has been introduced into molecular representation learning recently. In these works, a molecule is represented by either a SMILES \cite{doi:10.1021/ci00057a005} sequence or an undirected graph where the nodes and edges are atoms and bonds, respectively, By using SMILES sequence, \citet{wang2019smiles,chithrananda2020chemberta} use the masked language modeling objective~\citep{devlin2018bert} for pre-training. \citet{honda2019smiles} learn molecular representation by reconstructing the input SMILES with a Transformer based on a sequence-to-sequence model. By regarding a molecule as a graph, \citet{Hu2020Strategies} perform node-level and graph-level pre-training, which are about to reconstruct the masked attributes and preserve the consistency of similar subgraphs. \citet{wang2021molclr} use contrastive learning, where the representation of a molecule $m$ should be similar to the augmented version of $m$ while dissimilar to others.
GraphCL \cite{NEURIPS2020_3fe23034}  uses graph augmentation and contrastive learning for pre-training. \citet{rong2020grover} design GNN Transformer for pre-training, which extends the attention blocks to graph data. They also use two new training objective functions, one is to predict the motifs in a graph, and the other is to predict the subgraph property represented by a well-designed string. \citet{zhu2021dual} propose to maximize the consistency between SMILES representation and graph representation, and achieve remarkable performance on several downstream tasks.

\subsection{3D Molecular Representation}

Encoding 3D spatial structure into molecular representation is important to determine molecular property. \citet{schnet2017,lu2019molecular,anderson2019cormorant} take the atomic distance into consideration and design a set of novel architecture to deal with atomic positions.
\citet{klicpera_dimenet_2020,shui2020heterogeneous} 
further involve bond angle into their methods and achieve better performance. Recently, \citet{NEURIPS2020_15231a7c,PAINN_schutt21a} introduce equivariant networks to ensure the equivariance of molecular representation under continuous 3D roto-translations.

\subsection{Molecular Pre-training with 3D Information}
Recently, there emerge several works for molecule pre-training with 3D spatial structure. 
\citet{chen2021algebraic} use the element-specific
multiscale weighted colored algebraic graph (AG) to embed the chemical and physical interactions into graph invariants and capture 3D molecular structural information, which is then fused with the complementary bi-directional Transformer representation. 
\citet{stark20213d} propose to implicitly encode the 3D information into molecular representation by maximizing the mutual information between a 2D graph representation and a 3D representation which are produced by two separate networks. \citet{graphmvp} propose the Graph Multi-View Pre-training (GraphMVP) framework, which leverages constrastive learning and molecule reconstruction (under the variational auto-encoder framework) for pre-training.
To our best knowledge, almost all previous works encode 2D graph and 3D structure with different backbone models, and we are the first to encode them using a unified model.

\section{Our Method}
\subsection{Notations}
Let $G=(V,E)$ denote a 2D molecular graph, where $V = \{v_1, v_2,\cdots, v_{|V|}\}$ is a collection of atoms, and $E$ is a collection of bonds. Let $e_{ij}$ denote the bond between atom $v_i$ and $v_j$. 
Let $R\in\mathbb{R}^{|V|\times3}$ denote the 3D conformation of molecule $G$, where the $i$-th row $R$ is the coordinate of atom $v_i$.
For ease of reference, when the context is clear, we use $V$ to denote the indices of atoms, i.e., $V=\{1,2,\cdots,|V|\}$.

Let $\texttt{FF}(\cdots)$ denote a two-layer feed-forward network with ReLU activation and Batch Normalization~\citep{pmlr-v37-ioffe15}. The input of $\texttt{FF}(\cdots)$ is the concatenation of all input tensors, and the output is a $d$-dimensional vector, where $d$ is the dimension of the network hidden states.

\begin{figure*}[!htbp]
\centering
\includegraphics[width=0.8\linewidth]{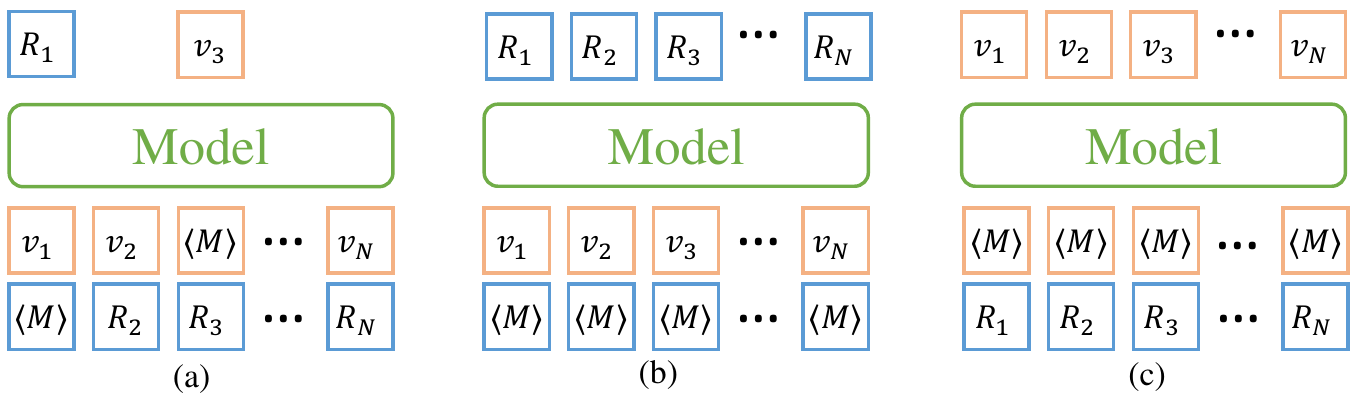}
\caption{The training objective functions. (a) reconstruction of masked atoms and coordinates; (b) 3D conformation generation conditioned on 2D graph; (c) 2D graph generation conditioned on 3D conformation.
 $\langle M\rangle$ denotes ths masked atoms and coordinates.}
\label{fig:overall_loss_function}
\end{figure*}

\subsection{Training Objective}

The overall training objective functions consist of three parts: reconstruction of masked atoms and coordinates; 3D conformation generation conditioned on 2D graph and 2D graph generation conditioned on 3D conformation. The illustration is in Figure \ref{fig:overall_loss_function}.

\subsubsection{Reconstruction of masked atoms and coordinates}\label{sec:reconstruct_atoms_coord}

~

Given a molecule $G$ with $|V|$ nodes and the corresponding conformation $R$, with probability $p\in(0,1)$, each atom $v_i$ is independently masked. Also, the coordinate of each atom $v_j$, i.e., $R_j$, is independently masked. Denote the indices of the unmasked atoms and coordinates as $I_{a}$ and $I_{c}$, respectively. The task is to reconstruct the masked tokens (i.e., $V\backslash{I_a}$ and $V\backslash{I_c}$) based on unmasked ones (i.e., $I_a$ and $I_c$). The training objective  for reconstructing the masked atoms is:
\begin{align}
& \ell_{\rm atom}=\sum_{i\in V\backslash{I_{a}}}\log P(v_i|\{v_j\}_{j\in I_a}; \{R_j\}_{j\in I_c}),\label{eq:lossFunction_atom_mlm}
\end{align}
which is similar to that in natural language processing \citep{devlin2018bert,liu2019roberta} and computer vision \cite{dosovitskiy2021an}.

\begin{figure}
\centering
\begin{minipage}{0.5\linewidth}
\centering
\subfigure[Symmetric molecule]{
\includegraphics[width=\linewidth]{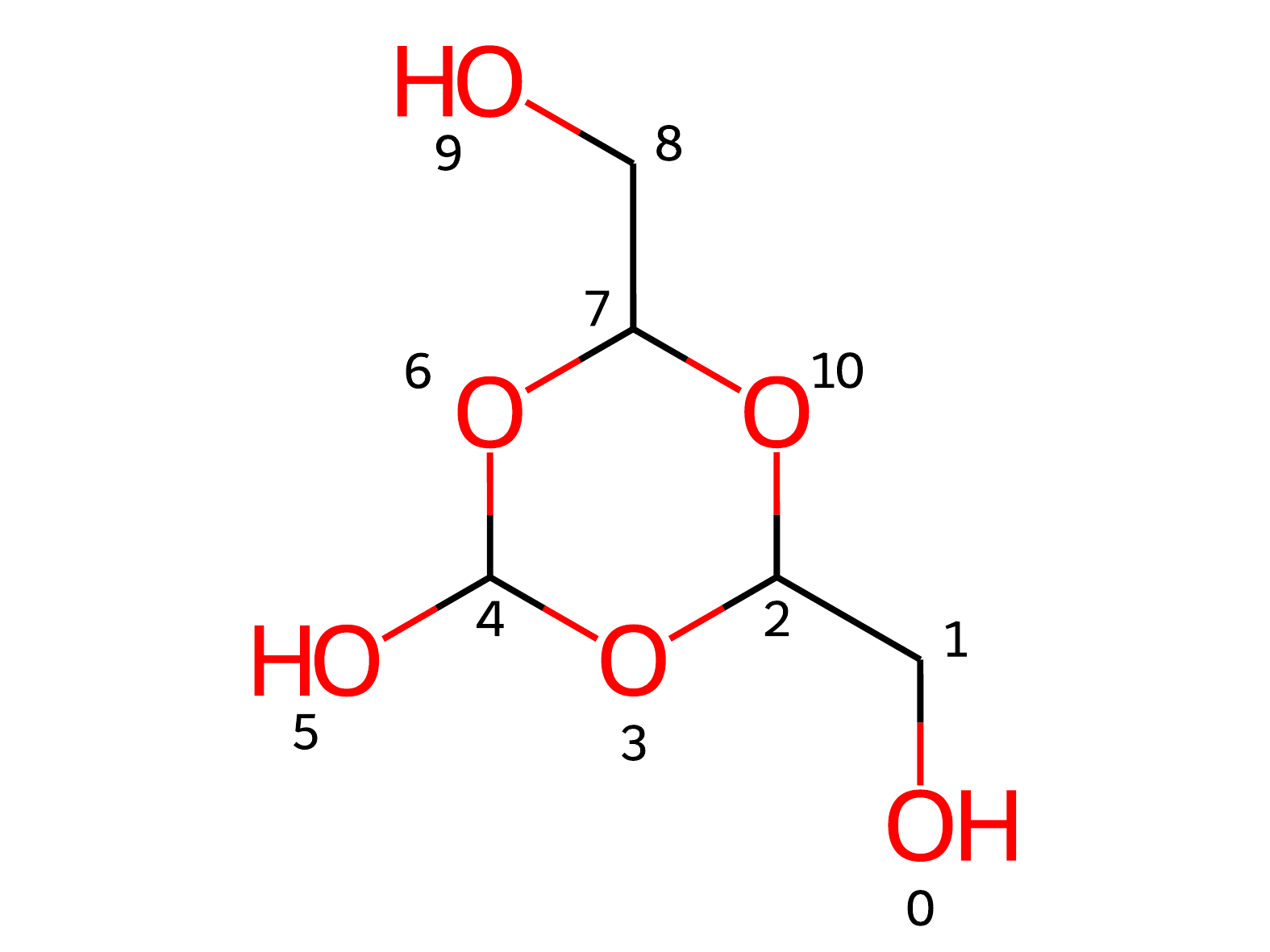}
}
\end{minipage}%
\begin{minipage}{0.5\linewidth}
\centering
\subfigure[Symmetric substructures.]{
\includegraphics[width=\linewidth]{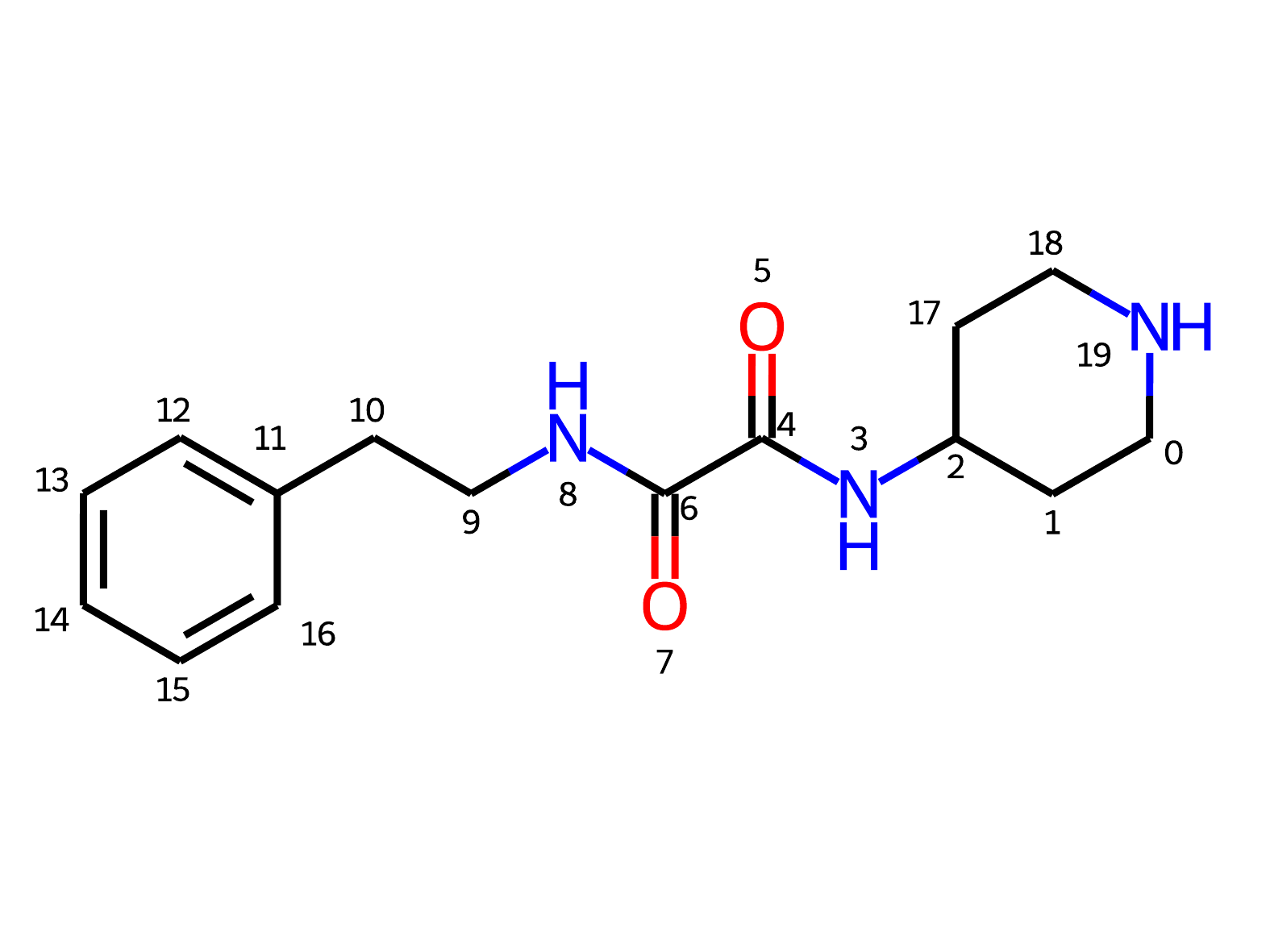}
}
\end{minipage}
\caption{An example of the symmetry in molecules.}
\label{fig:example_permutation}
\end{figure}

Assume the coordinate of atom $v_j$ is masked and let $\hat{R}_j$ denote the predicted coordinate of atom $v_j$. To measure the difference between the groundtruth coordinate $R_j$ and generated coordinate $\hat{R}_j$ $\forall j\in V\backslash{I_c}$, the most straightforward way is to use
\begin{equation}
\tilde{\ell}=\sum_{j\in V\backslash{I_c}}\Vert R_j-\hat{R}_j\Vert^2.
\label{eq:mse_loss_on_coord}
\end{equation}
However, for 3D conformation, this is not the best choice because there might be symmetric molecules, where the coordinates of atoms can be swapped. An example is shown in Figure~\ref{fig:example_permutation}(a), where the molecule is symmetric along the bond between atom $4$ and $5$. If we swap the coordinates of atom $3,2,1,0$ with $6,7,8,9$, the conformation remains the same, but the loss in Eqn.\eqref{eq:mse_loss_on_coord}  changes.
In addition, although some molecules are not symmetric, they still have symmetric substructures. As shown in Figure~\ref{fig:example_permutation}(b), atoms $12,13$ are symmetric to $16,15$ along the C-C bond (atom $10$ and $11$). Another symmetric substructure is the piperidine at the right side of Figure \ref{fig:example_permutation}(b). If we swap the coordinates of the symmetric substructures, the 3D conformation should remain unchanged. Such permutation invariance cannot be maintained by Eqn.\eqref{eq:mse_loss_on_coord}.

When we randomly mask the coordinates, it is possible to mask the symmetric substructures. To tackle this challenge, we follow \cite{Meli2020,zhu2022DMCG} and use the permutation invariant loss. Given a molecule $G$, let $\alpha$ denote a bijective mapping from $\{1,2,\cdots,|V|\}$ to itself. 
Let $\mathcal{P}$ denote the collection of atom mappings on symmetric substructures. For the picture in Figure \ref{fig:example_permutation}(b), $\mathcal{P}$ has four mappings.  

(1) $\alpha(i)=i$ $\forall i\in V$;

(2) $\alpha(12)=16,\alpha(16)=12,\alpha(13)=15,\alpha(15)=13$ and $\alpha(i)=i$ for remaining atoms;

(3) $\alpha(1)=17,\alpha(17)=1,\alpha(0)=18,\alpha(18)=0$ and $\alpha(i)=i$ for remaining atoms;

(4) $\alpha(12)=16,\alpha(16)=12,\alpha(13)=15,\alpha(15)=13$, $\alpha(1)=17,\alpha(17)=1,\alpha(0)=18,\alpha(18)=0,\alpha(i)=i$ for remaining atoms.

The loss function that is permutation invariant to symmetric substructures is defined as follows:
\begin{equation}
\ell_{\rm coord}=\min_{\alpha\in \mathcal{P}}\sum_{j\in V\backslash{I_{c}}}\Vert R_j-\alpha(\hat{R}_j)\Vert^2, \label{eq:lossFunction_coord_mlm}
\end{equation}
where $R_j\in\mathbb{R}^3$ and $\alpha(\hat{R})_j\in\mathbb{R}^3$ denote the coordinate of the $j$-th atom in $R$ and $\alpha(\hat{R})$, respectively. \citet{zhu2022DMCG} provide an efficient implementation to find the $\mathcal{P}$ based on graph automorphism and we follow their method.

\begin{figure*}
\centering
\includegraphics[width=0.9\linewidth]{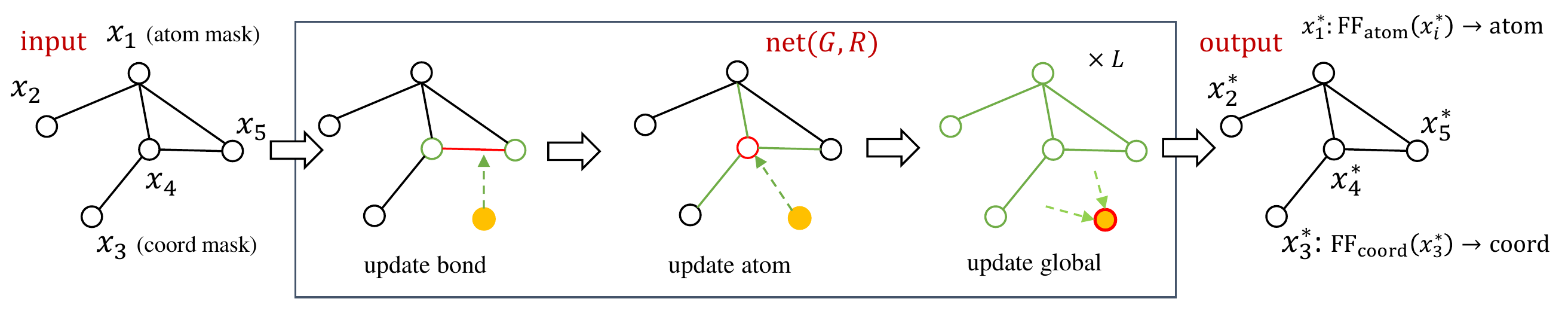}
\caption{A brief workflow of the network architecture. For the input molecule, we mask the atom information of atom $1$ and mask the coordinates of atom $3$. It is then processed by a stack of $L$ blocks, where the yellow node refers to the global representation of the molecule. We eventually obtain a representation $x^*_i$ for each atom, based on which we can reconstruct the masked atoms and coordinates.}
\label{fig:work_flow_netarch}
\end{figure*}

\subsubsection{3D conformation generation conditioned on 2D graph}\label{sec:3dgeneration}

~

We mask all the coordinates of a conformation and reconstruct them based on the 2D molecular graph only. Molecular conformation generation is an important topic for both bioinformatics and machine learning, and various methods have been proposed, including the distance-based methods \cite{pmlr-v119-simm20a,shi2021learning,xu2022geodiff} (which generate the interatomic distances or their gradients first and then reconstruct the conformation) or the direct approach (which directly outputs the 3D conformation). Recently, \citet{zhu2022DMCG} proposed a method named directly molecular conformation generation (DMCG) that outputs the coordinates directly and achieved state-of-the-art results. We use DMCG for conformation generation in our unified pre-training framework. 

The conformation should maintain roto-translation invariance \cite{Mansimov2019cvgae} and permutation invariance \cite{Meli2020}. Roto-translation invariance means that if we rotate and translate the generated 3D conformation $\hat{R}$ rigidly, the distance between $\hat{R}$ and the groundtruth conformation $R$ remains unchanged. The permutation invariance has been introduced in Section \ref{sec:reconstruct_atoms_coord}. Mathematically, the loss function is defined as follows:
\begin{equation}
\begin{aligned}
&\ell_\textrm{2D$\to$3D}=\frac{1}{|V|}\min_{\varrho,\alpha\in \mathcal{P}}\sum_{i\in V}\Vert R_i-\varrho(\alpha(\hat{R}_i))\Vert^2.\label{eq:lossFunction_2d-3d}
\end{aligned}
\end{equation}
In Eqn.\eqref{eq:lossFunction_2d-3d}, $\hat{R}_i$ is obtained based on the 2D graph $G$ without touching the 3D information; $\varrho$ is the roto-translational operation. $\varrho$  can be written as $\varrho(R)=RQ+t$, where $Q$ is a $3\times3$ rotation matrix, and $t\in R^{1\times3}$ is the transition vector. Eqn.\eqref{eq:lossFunction_2d-3d} is roto-translation invariant and permutation invariant to the symmetric substructures. According to \cite{karney2007quaternions}, calculating the minimum value over roto-translation operations can be transformed into calculating the minimum eigenvalues of some $4\times 4$ matrix, and we use their method. More details are available in our implementation.

\subsubsection{2D graph generation conditioned on 3D conformation}\label{sec:2dgenera}

~

We mask all atoms and reconstruct them based on the 3D conformation. The 2D structure of the molecule is kept, i.e., we know which two atoms are connected, but we do not know the bond type between them. The training objective function is defined as follows:
\begin{equation}
\ell_\textrm{3D$\to$2D}=\frac{1}{|V|}\sum_{i\in V}\log P(v_i|\{R_j\}_{j\in V}).
\label{eq:loss_3D-to-2D}
\end{equation}
Note that in Eqn.\eqref{eq:lossFunction_atom_mlm}, only partial atoms are masked. We can reconstruct them based on the remaining atoms and coordinates. In comparison,  Eqn.\eqref{eq:loss_3D-to-2D} is an extreme case where all atoms are masked, and we need to reconstruct them purely by coordinates. Both of the two loss functions are helpful for pre-training.

\subsubsection{Discussion}\label{sec:discussion}

~

Compared with the pre-training in natural language processing, $\ell_{\rm atom}$ and $\ell_{\rm 3D \to 2D}$ are similar to the masked language modeling \cite{devlin2018bert,liu2019roberta}. However, for $\ell_{\rm coord}$ and $\ell_{\rm 2D \to 3D}$, we use the permutation invariant and roto-translation invariant version. In Eqn.\eqref{eq:lossFunction_coord_mlm}, since only partial coordinates are masked, we just need to consider the permutation invariance of symmetric atoms and do not need the roto-translational operation on all coordinates. This is because the unmasked coordinates can help reduce the roto-translation freedom. In comparison, in Eqn.\eqref{eq:lossFunction_2d-3d}, considering all the atoms are masked, we should consider both permutation and roto-translation invariance since we have no unmasked coordinates to align.

\subsection{Network Architecture}\label{sec:net_arch}
As mentioned in Section~\ref{sec:intro}, we propose a method that can encode the 2D and 3D information using one model and output a unified representation. We use the GN block module proposed by \cite{deepmind2021ogb,zhu2022DMCG} as our backbone due to their superior performance on molecule classification and molecular conformation generation. Denote the model as  $\texttt{net}(G,R)$, where the inputs include the 2D graph $G$ and 3D conformation $R$. The input atoms, bonds are first mapped into $d$-dimensional representation using the corresponding embedding layers. All the masked atoms are represented by a special embedding $E_{a}$ and all the related bond embeddings are represented by another special embedding $E_b$, both of which are learned during pre-training.
If the coordinates $R_i\in\mathbb{R}^3$ of atom $i$ is masked, each element in $R_i$ is replaced with a number uniformly sampled from $[-1,1]$.

Compared with previous molecule pre-training with 2D information only, how to encode the 3D information is an important problem. Inspired by pointed Transformer \cite{zhao2021point} and the equivalent network \cite{NEURIPS2020_15231a7c}, we encode both the coordinates and interatomic distances. Let $x_i$ and $\bar{x}_i$ denote the representations of atom $i$ before and after combining with 3D information. Let $x_{ij}$ and $\bar{x}_{ij}$ denote the representations of bond $e_{ij}$ before and after combining with 3D information. We have that
\begin{equation}
\begin{aligned}
& \bar{x}_i = x_i + \texttt{FF}(R_i),\;
\bar{x}_{ij} = x_{ij} + \texttt{FF}(\Vert {R}_i- {R}_j \Vert).
\end{aligned}
\label{eq:embed_conformation_info}
\end{equation}
Note that some $x_i$ could be the $E_a$, some $x_{ij}$ could be $E_b$, and some coordinates $R_i$'s are not the real coordinates but randomly sampled. The $\bar{x}_i$'s and $\bar{x}_{ij}$'s are then fed into the main backbone for further processing. Briefly, \texttt{net}$(\cdots)$ stacks $L$ identical blocks. As shown in Figure \ref{fig:work_flow_netarch}, in each block, we first update the bond representations (the red edge) using the related atom representations (the green nodes) and a global representation of the molecule (the yellow node). After that, the atom representations (the red node) are updated in an attentive way using the related edge representations (the green edges) and the global representation (the yellow node). Finally, we update the global representation by averaging the updated atom and bond representations.
The network $\texttt{net}(G,R)$ eventually outputs a representation for each atom, i.e.,
\begin{equation}
(x^*_1,x^*_2\cdots,x^*_{|V|})=\texttt{net}(G,R).
\end{equation}
The mathematical formulation of $\texttt{net}(G,R)$ is summarized in Appendix \ref{app:network_details_app}.
We use two different MLP layers to
reconstruct the atoms and coordinates. Both of them take $x^*_i$ as input:

(1) $\texttt{FF}_{\rm atom}$, which outputs the masked atoms. It will be used in Eqn.\eqref{eq:lossFunction_atom_mlm} and Eqn.\eqref{eq:loss_3D-to-2D}.

(2)  $\texttt{FF}_{\rm coord}$, which outputs the masked coordinates. It will be used in Eqn.\eqref{eq:lossFunction_coord_mlm} and Eqn.\eqref{eq:lossFunction_2d-3d}.

For the example in Figure \ref{fig:work_flow_netarch}, we mask the atom information of node $x_1$ and mask the coordinates of node $x_3$. After obtaining their representations $x^*_1$ and $x^*_3$, they are reconstructed by $\texttt{FF}_{\rm atom}(x^*_1)$ and $\texttt{FF}_{\rm coord}(x^*_3)$

We summarize the pre-training procedure in Algorithm~\ref{alg:our_method}. 

\begin{algorithm}
\caption{Workflow of the pre-training.}
\label{alg:workflow}
\begin{algorithmic}[1]
\STATE \textit{Input}: Training data $(G,R)$; optimizer $\operatorname{opt}$; mask probability $p$;

{\it $\rhd$ Calculate $\ell_{\rm atom}$ and $\ell_{\rm coord}$ }

\STATE With probability $p$, we independently mask the atoms in $G$ and obtain $G'$. The unmasked atom indices are $I_a$.
\STATE With probability $p$, we independently replace the coordinates in $R$ with some coordinates uniformly sampled from $[-1,1]$ to obtain $R'$. The indices of the original coordinates are $I_c$.
\STATE $(\hat{x}^{*}_1,\hat{x}^{*}_2,\cdots,\hat{x}^*_{|V|})=\texttt{net}(G^\prime,R^\prime)$; 
\STATE If atom $v_i$ is masked, i.e., $i\in V\backslash{I_a}$,  $\hat{v}_i=\texttt{FF}_{\rm atom}(\hat{x}^*_i)$;
\STATE If coordinate $R_j$ is replaced, i.e., $j\in V\backslash{I_c}$,  $\hat{R}^\prime_j=\texttt{FF}_{\rm coord}(\hat{x}^*_j)$;
\STATE Calculate the $\ell_{\rm atom}$ in Eqn.\eqref{eq:lossFunction_atom_mlm} and the $\ell_{\rm coord}$ in Eqn.\eqref{eq:lossFunction_coord_mlm}.

{\it $\rhd$ Calculate $\ell_\textrm{2D$\to$3D}$ }
\STATE Replace all coordinates in $R$ with a random matrix $R^{\prime\prime}\in\mathbb{R}^{|V|\times3}$, where each element is uniformly sampled from $[-1,1]$; 
\STATE $(\tilde{x}^*_1,\tilde{x}^*_2,\cdots,\tilde{x}^*_{|V|})=\texttt{net}(G,R^{\prime\prime})$.
\STATE $\forall i\in V$, $\tilde{R}_i=\texttt{FF}_{\rm coord}(\tilde{x}^*_i)$.
\STATE Calculate the loss $\ell_\textrm{2D$\to$3D}$ in Eqn.\eqref{eq:lossFunction_2d-3d} using  $\{\tilde{R}_i\}_{i\in V}$.

{\it $\rhd$ Calculate $\ell_\textrm{3D$\to$2D}$ }
\STATE Mask all atoms and bonds in $G$ and obtain $G^{\prime\prime}$. The connection of atoms (i.e., which two atoms are connected) is kept.
\STATE $(\bar{x}^{*}_1,\bar{x}^{*}_2,\cdots,\bar{x}^{*}_{|V|})=\texttt{net}(G^{\prime\prime},R)$. 
\STATE $\forall i\in V$, $\bar{v}_i=\texttt{FF}_{\rm atom}(\bar{x}^{*}_i)$.
\STATE Calculate the loss $\ell_\textrm{3D$\to$2D}$ in Eqn.\eqref{eq:loss_3D-to-2D} using  $\{\bar{v}_i\}_{i\in V}$.

 {\it $\rhd$ Update the model parameter $\theta$. }
\STATE Denote the parameter of \texttt{net} as $\theta$. Update the model by

$\theta\leftarrow\operatorname{opt}(\theta,\nabla_\theta(\ell_{\rm atom} + \ell_{\rm coord} + \ell_\textrm{2D$\to$3D} + \ell_\textrm{3D$\to$2D}))$.
\end{algorithmic}
\label{alg:our_method}
\end{algorithm}

\begin{table*}[!htpb]
\centering
\caption{Results on MoleculeNet, where only 2D information is available.}
\begin{tabular}{lccccccc}
\toprule
Dataset& BBBP  & Tox21 &ClinTox & HIV&BACE&SIDER&Avg\\
\midrule
AttrMask \& ContextPred \cite{Hu2020Strategies} & $71.2\pm 0.9$ & $74.2\pm0.8$ & $73.7\pm 4.0$ & $75.8\pm 1.1$ & $78.6\pm1.4$ & $60.4\pm0.6$ & $72.31$ \\
GROVER \cite{rong2020grover} & $70.3\pm1.6$ & $75.2\pm0.3$ & $77.8\pm2.0$ & $75.9\pm0.9$ & $79.2\pm0.3$ & $60.6\pm 1.1$ & $73.17$\\
GraphCL \cite{NEURIPS2020_3fe23034} & $67.5\pm3.3$ & $75.0\pm0.3$ & $78.9\pm4.2$ & $75.0\pm0.4$ & $68.7\pm7.8$ & $60.1\pm1.3$ & $70.87$ \\
\citet{stark20213d} & $71.1\pm2.0$&$78.9\pm0.6$&$59.4\pm3.2$&$76.1\pm1.1$&$79.4\pm2.0$&$57.3\pm5.0$ & $70.37$\\
GraphMVP \citep{graphmvp}&$72.4\pm1.6$&$75.9\pm0.5$&$77.5\pm4.2$&$77.0\pm1.2$&$81.2\pm0.9$&$63.9\pm1.2$&74.65\\
\midrule
Ours &$77.4\pm0.6$& $75.9\pm0.3$&$95.4\pm1.1$&$82.2\pm1.0$&$86.8\pm0.6$&$67.4\pm0.5$ & $80.85$\\
\bottomrule
\end{tabular}
\label{tab:moleculenet}
\end{table*}

\begin{table}[!h]
\centering
\caption{Results on ogb-molpcba dataset.}
\begin{tabular}{lcc}
\toprule
Method & Valid AP & Test AP \\
\midrule
GN block~\citep{deepmind2021ogb} & 0.2745&0.2650 \\
GIN + virtual node \citep{kong2021flag_paper} & 0.2798&0.2703\\
Graphormer~\citep{ying2021graphormer} & 0.3227 & 0.3140\\
\midrule
Ours & 0.3225 & 0.3125 \\
Ours + FLAG & 0.3304 & 0.3174 \\
\bottomrule
\end{tabular}
\label{tab:main_results_molpcba}
\end{table}

\begin{table*}[!htbp]
\centering
\caption{Results of toxicity prediction where both 2D and 3D information is available.}
\resizebox{\linewidth}{!}{
\begin{tabular}{lcccccccccccc}
\toprule
& \multicolumn{3}{c}{LD50} & \multicolumn{3}{c}{LC50} & \multicolumn{3}{c}{IGC50} & \multicolumn{3}{c}{LC50DM} \\
& $R^2$ ($\uparrow$) & RMSE ($\downarrow$) &MAE  ($\downarrow$) & $R^2$($\uparrow$) & RMSE ($\downarrow$)&MAE ($\downarrow$) & $R^2$ ($\uparrow$) & RMSE ($\downarrow$)&MAE ($\downarrow$) & $R^2$($\uparrow$) & RMSE  ($\downarrow$)&MAE ($\downarrow$)\\
\midrule
MACCS   &  0.643 & N/A & N/A&  0.608 & N/A& N/A & 0.643 & N/A & N/A & 0.434  & N/A& N/A\\ 
Daylight & 0.624 & N/A &N/A&  0.724 & N/A &N/A & 0.717 & N/A& N/A& 0.700 & N/A&N/A \\
3D MT-DNN & 0.653 &  0.568 & 0.421 & 0.789 & 0.677& 0.446 & 0.802 &  0.438 & 0.305 & 0.678 & 0.978 & 0.714 \\
AGBT & 0.671 & 0.554&0.401 & 0.783 & 0.692&0.492 & 0.842 &  0.391&0.273 & 0.830 & 0.743&0.527 \\
Ours & 0.690 & 0.542 &0.399& 0.800 & 0.664&0.444 & 0.858 & 0.379 &0.263& 0.833 & 0.715 &0.475\\
\bottomrule
\end{tabular}}
\label{tab:results_toxicity_2d_3d}
\end{table*}

Please note that \cite{zhu2022DMCG} is for 2D molecular graph to 3D conformation generation, and their loss function is specially designed for conformation generation.
In this work, our focus is to jointly use 2D/3D information and obtain effective molecular representations.
Our method can be used for molecule modeling and generation.

\section{Application to Property Prediction}
We work on $11$ tasks to verify the effectiveness of our method: (1) $7$ molecular property prediction tasks with 2D information only; (2) $4$ molecular property prediction tasks with both 2D and 3D information.

\subsection{Settings}
\paragraph{Dataset} We use the PCQM4Mv2 dataset~\citep{hu2021ogblsc} for pre-training, which has $3.38$M data. In PCQM4Mv2, both the 2D information and 3D information are available. In this dataset, each 2D molecular graph corresponds to one 3D conformation calculate by density function theory. We randomly split the dataset into training and validation sets by $95\%$:$5\%$. 

For molecular property prediction with 2D information only, following \cite{graphmvp,stark20213d}, we choose six tasks on MoleculeNet \cite{Wu2017moleculenet}: BBBP, Tox21, ClinTox, HIV, BACE and SIDER. Most of them are with limited data. Following \cite{graphmvp,stark20213d}, for each task, we split the dataset into training, validation and test sets by 8:1:1 according to their molecular scaffolds. We also conduct experiments on ogb-molpcba~\citep{ogb-smallscale}, which is a larger dataset with $438$K  data. All these tasks are classification tasks.

For property prediction with both 2D and 3D information, following \citet{chen2021algebraic}, we work on four toxicity prediction tasks: LD50, IGC50, LC50 and LC50DM, all of which are regression tasks. They measure the toxicity from different aspects. More details of the above datasets are summarized in Table \ref{tab:dataset_description} of the Appendix.

\paragraph{Training configuration} The pre-trained model has 12 layers, and the hidden dimension of each block is $256$. The masked ratio $p$ is $0.25$. The model is pre-trained for $100$ epochs using Adam optimizer with initial learning rate $2\times10^{-4}$, batch size $128$ and is trained on four P40 GPUs.

For the molecular property prediction tasks with 2D information only, all the coordinates are randomly sampled from uniform distribution $[-1,1]$. For the tasks on MoleculeNet, we use grid search to determine the learning rate, dropout and batch size, which are summarized in Appendix \ref{sec:morehyperparams}. For ogb-molpcba, the learning rate is fixed as $10^{-4}$ and we train the model for $100$ epochs. 

For the four toxicity prediction tasks with 3D information, following \cite{chen2021algebraic}, we use multitask learning to jointly tune the four tasks. We also use grid search to find the hyper-parameters, which are left in Appendix \ref{sec:morehyperparams}.

\paragraph{Evaluation} For MoleculeNet, we use area under the receiver operating characteristic curve (ROC-AUC). Each experiment is independently run for three times with different seeds, and the mean and standard derivation are reported. Ogb-molpcba has $128$ sub-tasks and we use average precision (briefly, AP) as suggested by \citet{ogb-smallscale}. For the four toxicity prediction, following \citep{chen2021algebraic},  we use squared Pearson correlation coefficient ($R^2$) between the groundtruth and predicted values, rooted mean square error (RMSE) and mean absolute error (MAE) as the evaluation metrics.

\paragraph{Baselines} For MoleculeNet, we compare with three 2D pre-training baselines:  AttrMask \& ContextPred \cite{Hu2020Strategies},  GROVER \cite{rong2020grover} and GraphCL \cite{NEURIPS2020_3fe23034}. They are all introduced in related work. We also compare with two 3D pre-training baselines: \citet{stark20213d} and GraphMVP \citep{graphmvp}, which have been discussed comprehensively. For ogb-molpcba, we mainly compare with Graphormer \cite{ying2021graphormer}, which is also pre-trained on PCQM4M and achieves the best result on this dataset. For the four toxicity prediction tasks where 3D information is available, we mainly compare with AGBT \cite{chen2021algebraic}. Since the Github repository of GraphMVP is empty by the end of the submission, we leave the comparison with it on the four toxicity prediction tasks in the future.
We also compare with some classical baselines, which mainly use various molecular fingerprints and multi-task learning.

\begin{figure*}
\centering
\begin{minipage}{0.3\linewidth}
\subfigure[BBBP]{
\centering
\includegraphics[width=\textwidth]{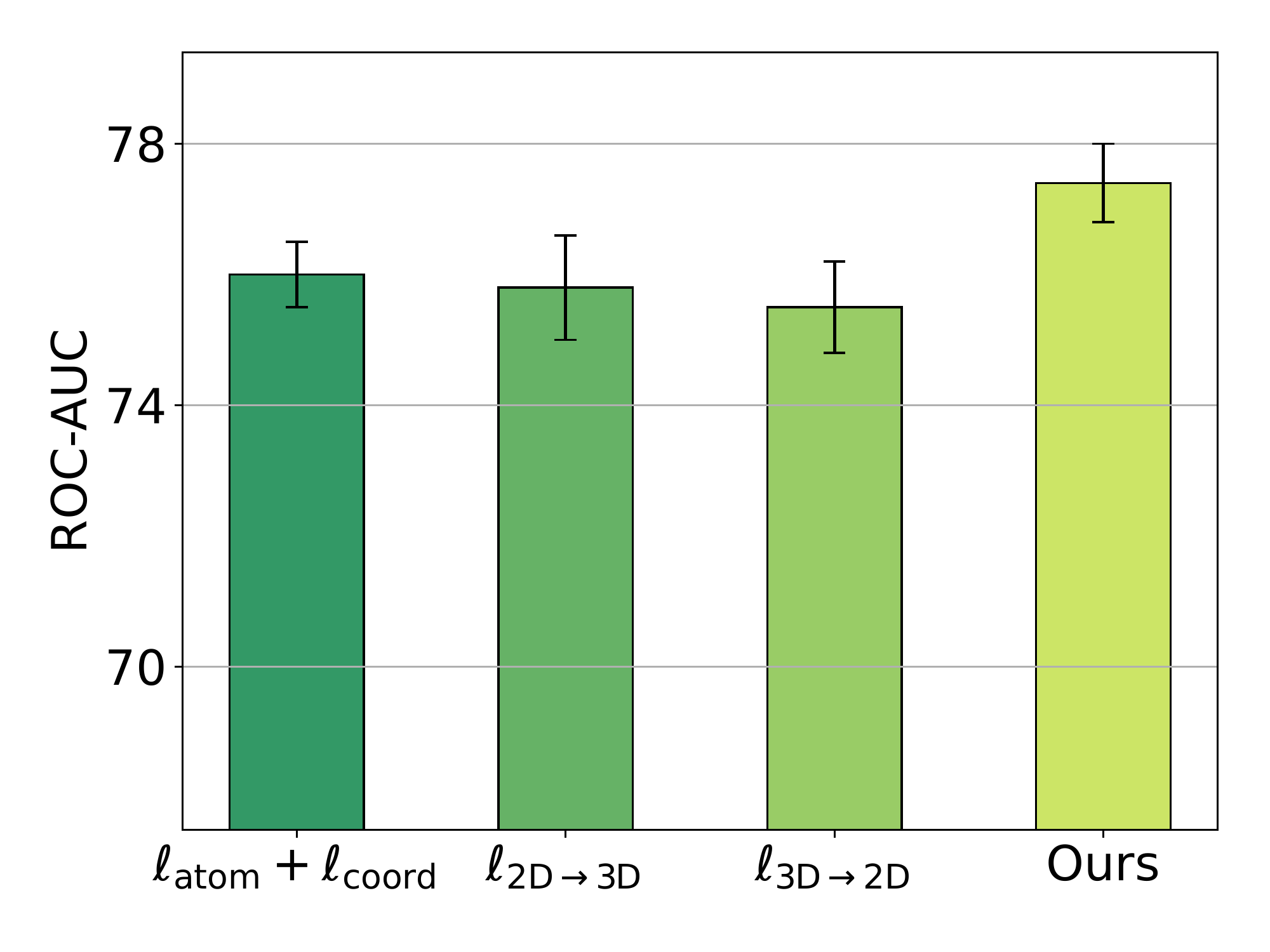}
}
\end{minipage}
\begin{minipage}{0.3\linewidth}
\subfigure[ClinTox]{
\centering
\includegraphics[width=\textwidth]{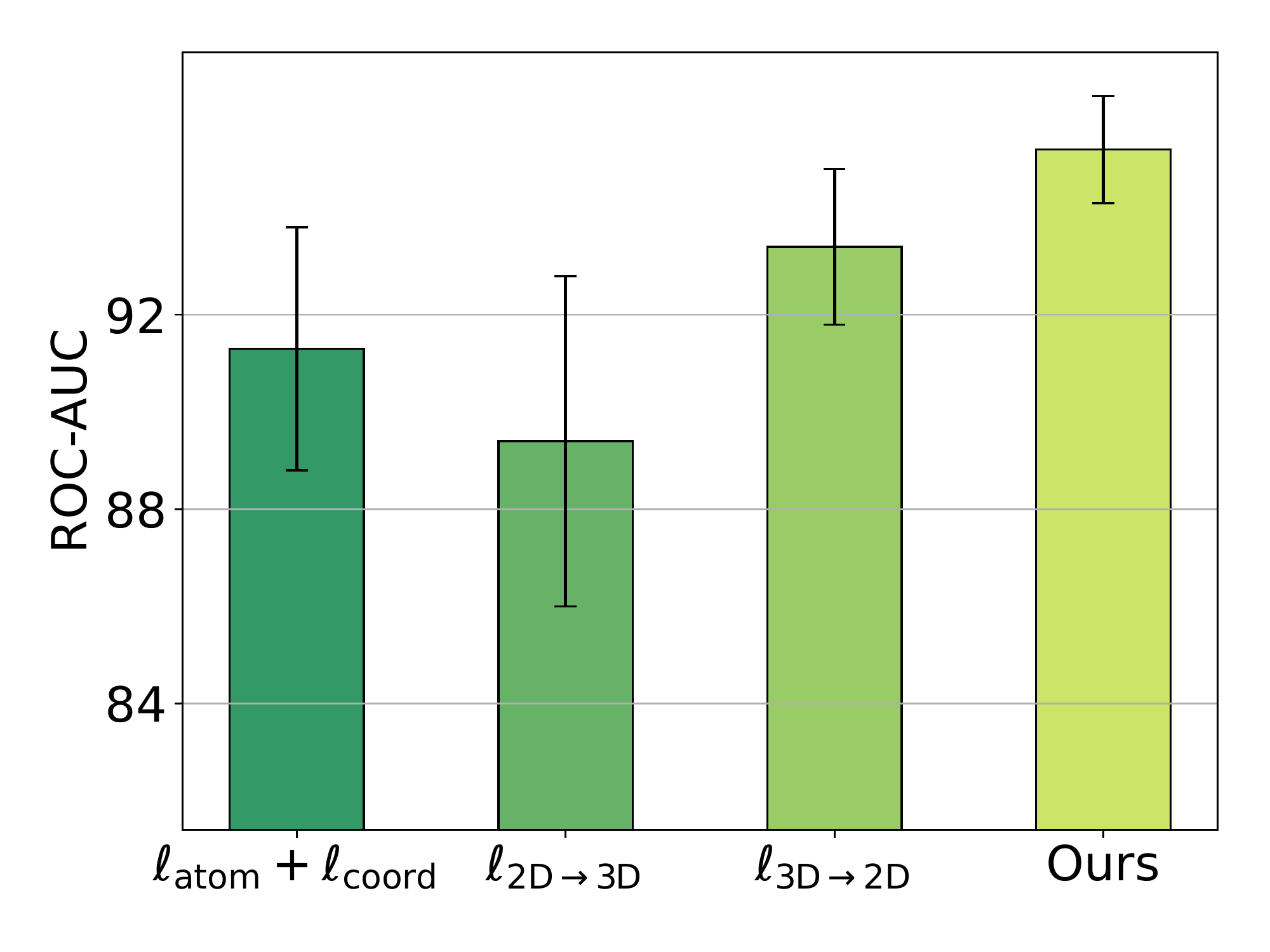}
}
\end{minipage}
\begin{minipage}{0.3\linewidth}
\subfigure[ogb-molpcba]{
\centering
\includegraphics[width=\textwidth]{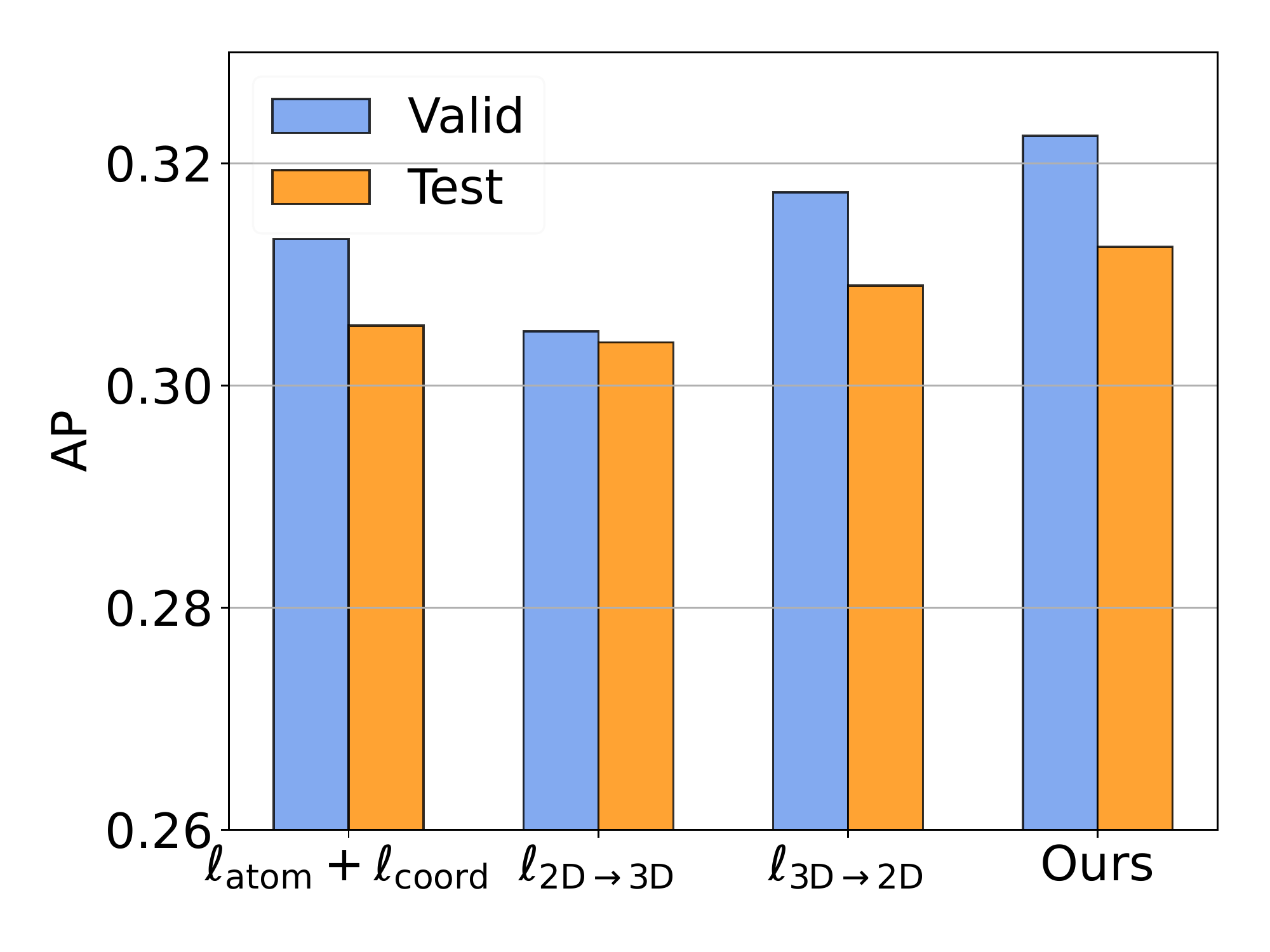}
}
\end{minipage}
\caption{Comparison of different pre-training objective functions.}
\label{fig:diff_pretrain_loss_function_property_pred}
\end{figure*}

\subsection{Results}
The results on MoleculeNet are reported in Table \ref{tab:moleculenet}. We have the following observations:

(1) Compared with AttrMask \& ContextPred \cite{Hu2020Strategies}, GROVER \cite{rong2020grover} and GraphCL \cite{NEURIPS2020_3fe23034} which only leverage the 2D molecular graphs for pre-training, our method outperforms these baselines by large margins. Specifically, on average, our method improves those baselines by $11.81\%$, $10.50\%$ and $14.08\%$. This shows the effectiveness of using 3D information in pre-training.

(2) Compared with \citet{stark20213d} and GraphMVP \cite{graphmvp}, our method still outperforms those baselines. We can improve the average ROC-AUC from $70.37$ and $74.65$ to $80.85$. \citet{stark20213d} and \citet{graphmvp} use two separated modules to encode the 2D and 3D information, and they use consistency loss to build their connection. In comparison, we use a unified module to deeply fuse them. The improvement shows that using unified representation is a promising direction for molecule modeling.

The results on ogb-molpcba are summarized in Table~\ref{tab:main_results_molpcba}. We report the results on both validation and test sets following the guidelines \cite{ogb-smallscale}. GN \cite{deepmind2021ogb} and ``GIN + virtual node'' \cite{kong2021flag_paper} are two baselines without pre-training, where ``GN'' is the backbone of our model, and ``GIN + virtual node'' denotes applying a virtual node, which aggregates the global information of a molecule, into the GIN model \cite{xu2019GIN}. The Graphormer \cite{ying2021graphormer} here  uses an adversarial augmentation technique named FLAG \cite{kong2021flag_paper} and obtains good improvements to the non-FLAG version. We also combine FLAG with our method in this work.

We can see that, although ogb-molpcba has  more data, it still benefits from pre-training. Compared with GN block, which is the non-pretraining version of network architecture, we can improve the test score from $0.2650$ to $0.3125$. In addition, by using FLAG, our method outperforms Graphormer by $2.4\%$ and $1.1\%$ on validation and test sets, setting new records for this task.

The results of toxicity prediction with 3D information are reported in Table \ref{tab:results_toxicity_2d_3d}. Compared with AGBT \cite{chen2021algebraic}, our method outperforms this strong baseline on all the four tasks and evaluation metrics. \citet{chen2021algebraic} separately deal with the 2D graph and 3D conformation, while we use a unified model to extract representations of the input molecule that can more consistently utilize the two types of information.
Our method also outperforms the non-pretraining methods, including the 2D fingerprint baselines like MACCS \cite{doi:10.1021/ci010132r}, Daylight(reported by \cite{D0CP00305K}), 
and the 3D based methods like 3D MT-DNN \citep{doi:10.1021/acs.jcim.7b00558}. This shows that the molecular representation obtained by pre-training outperforms the manually designed fingerprints. These results demonstrate the effectiveness of our proposed method for molecular property prediction.


\subsection{Ablation Study}\label{sec:ablation}

We mainly focus on the following questions:

\noindent(Q1) What is the contribution of each training objective function?

\noindent(Q2) What is the effect of different mask ratios?

To answer (Q1), we independently pre-train the models with $\ell_{\rm atom} + \ell_{\rm coord}$, $\ell_\textrm{2D$\to$3D}$ and $\ell_\textrm{3D$\to$2D}$, respectively. After that, we finetune the obtained models on the MoleculeNet datasets and ogb-molpcba. Part of the results are summarized in Figure \ref{fig:diff_pretrain_loss_function_property_pred} and the remaining ones are left in Appendix \ref{sec:diff_loss}. We can see that: 

(1) If we independently use any one of $\ell_{\rm atom} + \ell_{\rm coord}$, $\ell_\textrm{2D$\to$3D}$ and $\ell_\textrm{3D$\to$2D}$, the performance is not as good as using them together. This shows that each of them contribute to the unified pre-training and we should combine them together.

(2) Reconstructing the masked atoms and coordinates is relatively more important for the unified pre-training. However, it is ignored by previous works  \cite{graphmvp,stark20213d}, which mainly focus on preserving the consistency of different views. 

(3) According to our statistics, the avarage ROC-AUC scores on MoleculeNet of the above three loss functions are $79.15$, $78.57$ and $79.21$. All of them are better than \cite{graphmvp,stark20213d}, which demonstrate the power of unified representations.

\begin{figure}[!htbp]
    \small 
    \centering
    \begin{minipage}{0.48\linewidth}
    \subfigure[BBBP]{
    \includegraphics[width=\textwidth]{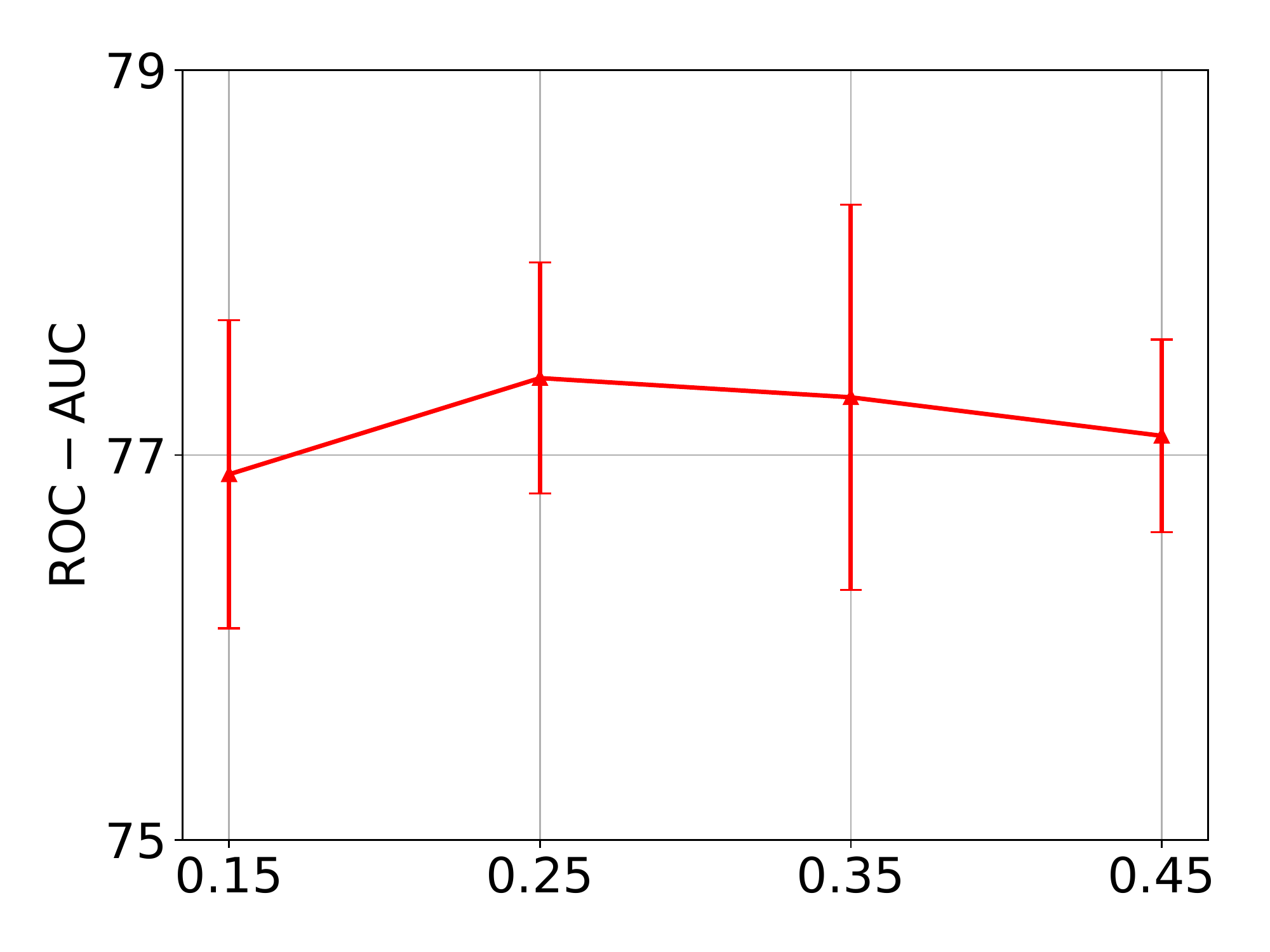}}
    \end{minipage}
    \begin{minipage}{0.48\linewidth}
    \subfigure[ClinTox]{
    \includegraphics[width=\textwidth]{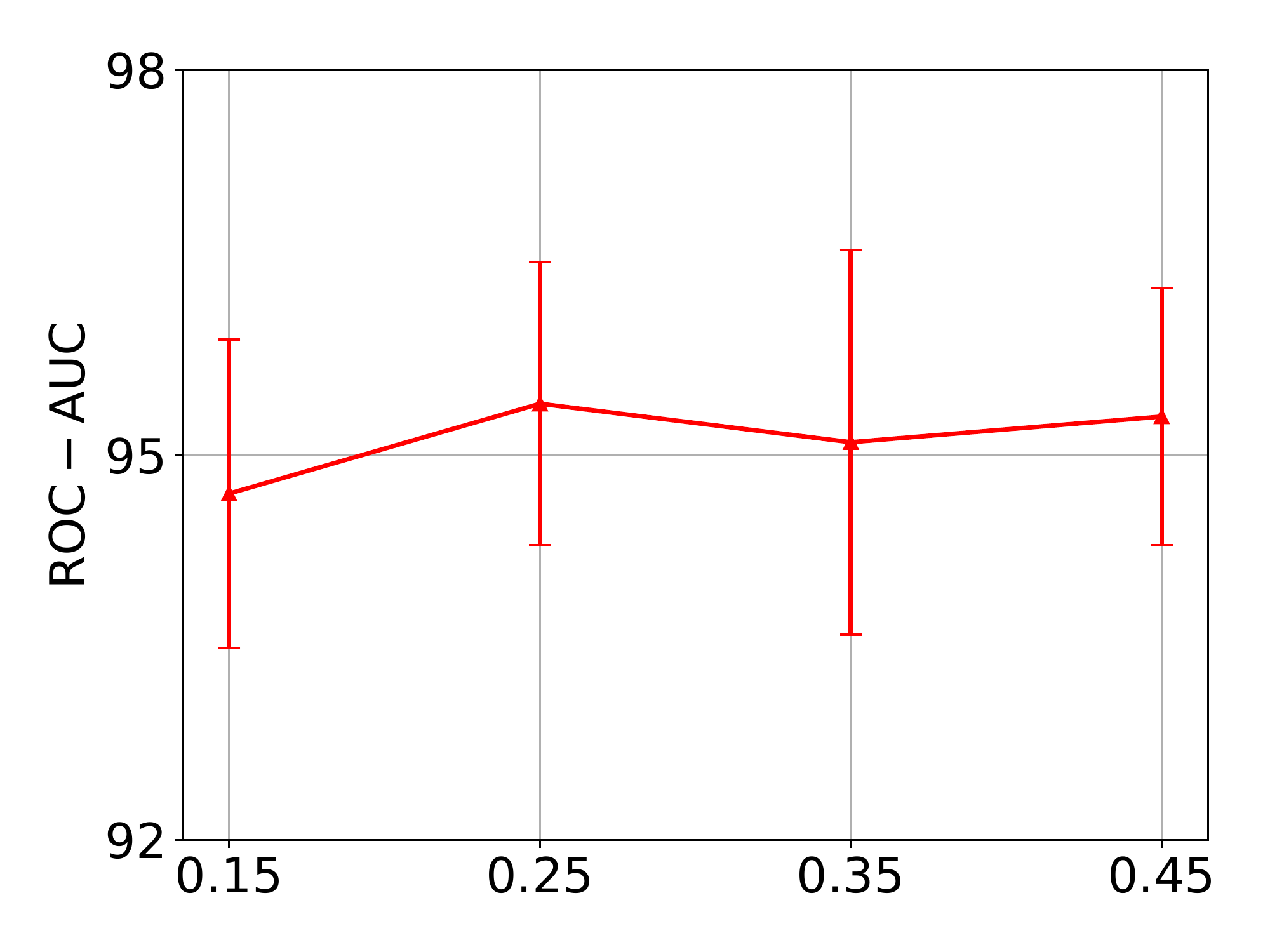}}
    \end{minipage}
    \caption{Results of different masked ratios $p$.}
    \label{fig:ablation_diff_ratio}
\end{figure}

To answer (Q2), we try 4 different mask ratios, $0.15$, $0.25$, $0.35$ and $0.45$. We conduct experiments on the BBBP and Clintox from MoleculeNet. The results are shown in Figure \ref{fig:ablation_diff_ratio}. In general, our pre-training method is robust to the choice of the mask ratios. For example, on BBBP, the maximum performance gap among the four ratios is $0.5$. Empirically, setting $p$  as $0.25$ achieves the best results.

\begin{table*}[!htb]
\small
\centering
\caption{Experimental results on conformation generation. }
\begin{tabular}{l cccc c cccc}
\toprule
Dataset&\multicolumn{4}{c}{QM9}&& \multicolumn{4}{c}{Drugs}\\
\multirow{2}{*}{Methods} &\multicolumn{2}{c}{ COV(\%)$\uparrow$ } & \multicolumn{2}{c}{MAT (\AA)$\downarrow$}  && \multicolumn{2}{c}{ COV(\%)$\uparrow$ } & \multicolumn{2}{c}{MAT (\AA)$\downarrow$} \\ 
&Mean &Median&Mean &Median&&Mean &Median&Mean &Median\\
\midrule
RDKit& 83.26& 90.78& 0.3447& 0.2935&&60.91& 65.70& 1.2026& 1.1252\\
GraphDG \cite{pmlr-v119-simm20a} & 73.33 & 84.21 & 0.4245 & 0.3973 & & 8.27 & 0.00 & 1.9722 & 1.9845 \\
ConfGF \cite{shi2021learning} & ${88.49}$ &94.13& ${0.2673}$& 0.2685&& 62.15 &70.93& 1.1629& 1.1596\\
DGSM \cite{luo2021DGSM} & 91.49 & 95.92 & 0.2139 & 0.2137 && 78.73 & 94.39 & 1.0154 & 0.9980 \\
GeoDiff \cite{xu2022geodiff} & 91.68 & 95.82 & 0.2099 &  0.2021 & & 89.13 & 97.88 & 0.8629 & 0.8529 \\
         DMCG \cite{zhu2021dual} &${96.34}$&${99.53}$&${0.2065}$& ${0.2003}$ &&${96.69}$&${100.00}$&${0.7223}$&${0.7236}$\\
         \midrule
         Ours &96.93&100.00& 0.1958&0.1849&&97.05&100.00&0.7056&0.6973\\
         \bottomrule
\end{tabular}
\label{tab:molecular_conf_gen}
\end{table*}

When only 2D information is available, people might be curious about the initialization of the 3D coordinates. By default, we uniformly sample the coordinates from $[-1,1]$. To verify its robustness, on ogb-molpcba, we randomly sample ten groups of initial coordinates and evaluate them. We also set all the initial coordinates as zero and check the inference result. The differences of the above $11$ trials are less than $10^{-4}$, which shows that our model is robust to the initial coordinates.

\section{Application to  Conformation Generation}
In this section, we conduct experiments on 3D conformation generation, which is to map the 2D molecular graph to 3D conformation.

\subsection{Settings}
Following~\citep{pmlr-v119-simm20a,xu2022geodiff,luo2021DGSM}, we use a subset of GEOM-QM9 and GEOM-Drugs \cite{axelrod2020geom} to evaluate our method. The numbers of training, validation and test (molecule, conformation) pairs of GEOM-QM9 are $200$K, $2.5$K and $22408$ respectively, while those numbers for GEOM-Drugs are $200$K, $2.5$K and $14324$. On average, GEOM-QM9 and GEOM-Drugs have 8.8 and 24.9 heavy atoms respectively.
 
Considering each molecular corresponds to multiple diverse conformations, we adopt the variational auto-encoder framework and generate multiple conformations. We choose the model proposed by \citet{zhu2022DMCG} for conformation generation, and the parameters are initialized by our pre-trained unified model. The model is a $12$-layer network with hidden dimension $256$. We also use Adam optimizer for training with learning rate $10^{-4}$ and the model is finetuned for $100$ epochs.
 
We evaluate the diversity and accuracy of generated conformations. Let $\operatorname{RMSD}(R,\hat{R})$ denote the root mean square deviations of two conformations $R$ and $\hat{R}$:
\begin{small}
\begin{equation}
\operatorname{RMSD}(R,\hat{R}) = \min_{\Phi}\left(\frac{1}{|V|}\sum_{i=1}^{|V|}
\Vert R_i-\Phi(\hat{R}_i)\Vert^2\right)^{\frac{1}{2}},
\end{equation}
\end{small}
where $\Phi$ is the alignment function that aligns two conformations by roto-translation operations. We use the coverage score (COV) and matching score (MAT) for evaluation. Let $S_g$ and $S_r$ denote the collections of the generated conformations and the groundtruth conformations. Assume in the test set, the molecule $m$ has $N_m$ conformations, following \cite{shi2021learning,luo2021DGSM}, we generate $2N_m$ conformations for it. Mathematically, COV$(S_g,S_r)$ and MAT$(S_g,S_r)$ are defined as follows:
\begin{small}
\begin{equation}
\begin{aligned}
& \operatorname{COV}(S_g,S_r)=\frac{1}{|S_r|}\big|\{R\in S_r|\operatorname{RMSD}(R,\hat{R})<\delta,\exists\hat{R}\in S_g \}\big|;\\
& \operatorname{MAT}(S_g,S_r)=\frac{1}{|S_r|}\sum_{R\in S_r}\min_{\hat{R}\in S_g}\operatorname{RMSD}(R,\hat{R}).
\end{aligned}
\end{equation}
\end{small}
Following previous works \cite{shi2021learning,xu2022geodiff}, $\delta$'s are set as 0.5 and 1.25 for GEOM-QM9 and GEOM-Drugs datasets, respectively.

\begin{figure}
\small
    \centering
    \begin{minipage}{0.53\linewidth}
    \subfigure[MAT scores  ($\downarrow$). ]{
    \includegraphics[width=\textwidth]{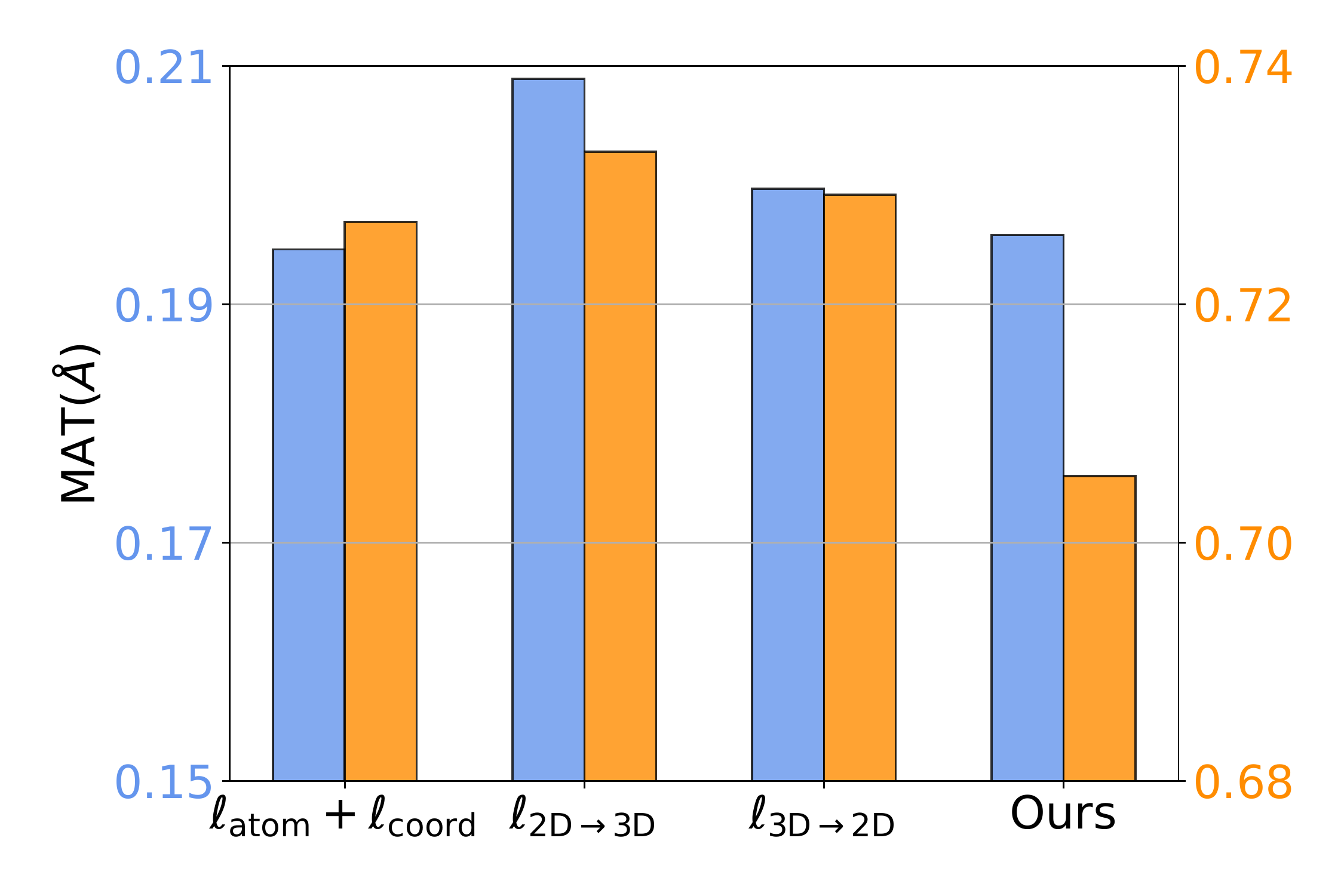}}
    \end{minipage}
    \begin{minipage}{0.458\linewidth}
    \subfigure[COV scores ($\uparrow$).]{
    \includegraphics[width=\textwidth]{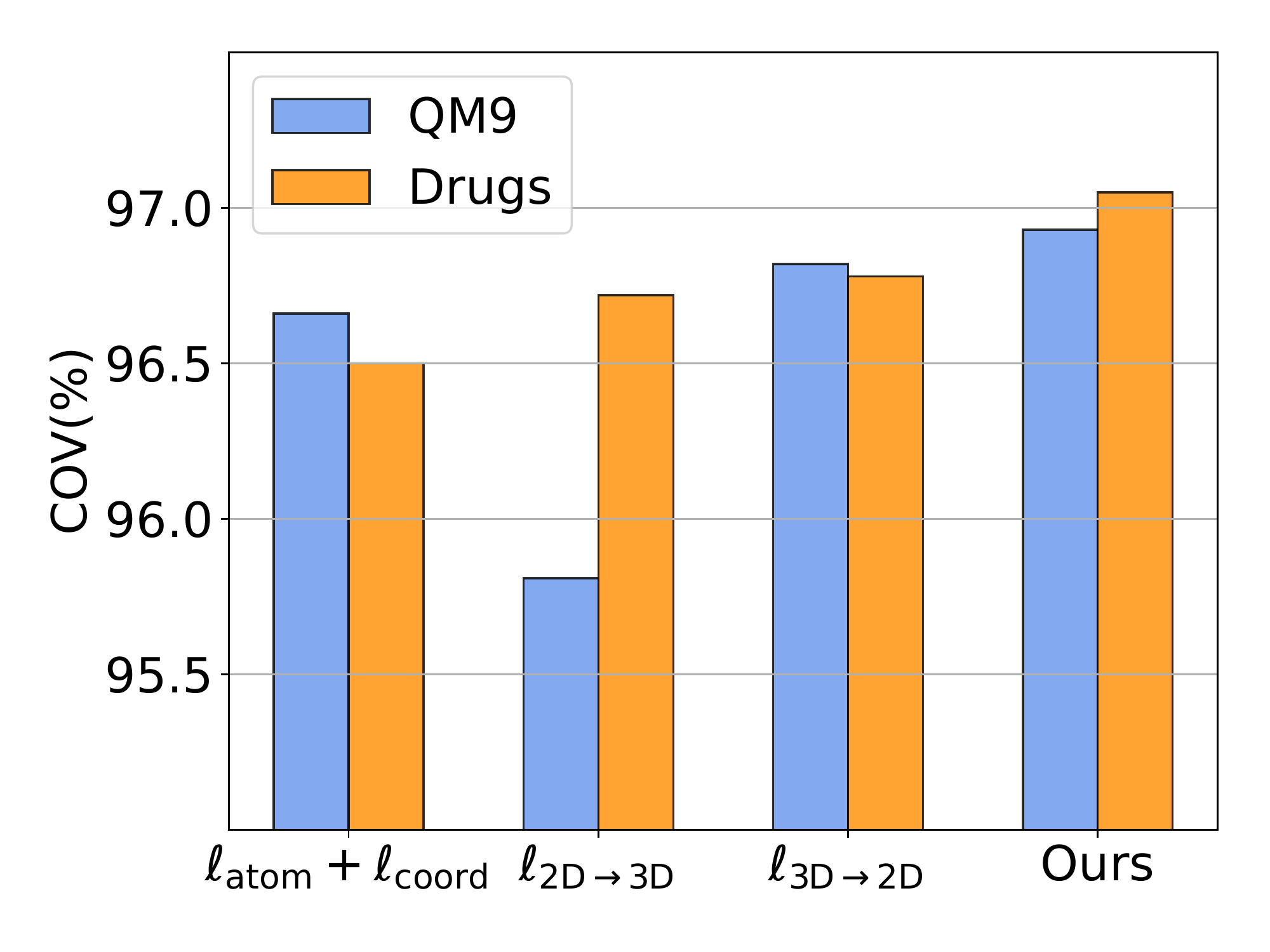}}
    \end{minipage}
    \caption{ Comparison of different pre-training objective functions on conformation generation tasks.}
    \label{fig:confgen_diff_loss}
\end{figure}

\subsection{Results}
The results are reported in Table \ref{tab:molecular_conf_gen}. We can see that our method outperforms all previous baselines, no matter for distance-based methods like ConfGF~ \cite{shi2021learning}, DGSM~\cite{luo2021DGSM},  or the recently proposed direct approach~\cite{zhu2022DMCG}. This shows the effectiveness of our pre-training in conformation generation. 

Similar to the first question in Section \ref{sec:ablation}, we also conduct ablation study of different training objective functions. The results are shown in Figure \ref{fig:confgen_diff_loss}. For the MAT scores, the values of QM9 correspond to the left axis, and those of Drugs correspond ot the right axis. We get the same conclusion as Section \ref{sec:ablation}: Using all the three types of loss function together achieves the best performance.


\section{Conclusions and Future Work}
In this work, we propose the unified 2D and 3D pre-training, that jointly leverages the 2D graph structure of the molecule and the 3D conformation. We design three loss functions, including reconstructing the masked atoms and coordinates, 2D graph to 3D conformation generation and the 3D conformation to 2D graph generation. The permutation invariance and roto-translation invariance are  considered. We conduct experiments on 11 molecular property prediction tasks and 2 conformation generation tasks, and achieve state-of-the-art results on 12 of them.

For future work, first, we will explore how to leverage the large amount of molecules without 3D structures. We could use them to enhance the $\ell_{\rm atom}$ and use cycle consistency loss, where we  map the 2D graph $G$ to 3D conformation $\hat{R}$ and then map $\hat{R}$ back to 2D graph $\hat{G}$. $G$ and $\hat{G}$ should preserve enough similarity. Second, the combination of our method with contrastive learning is another topic. Third, we will apply our pre-trained model to more scenarios like drug-drug interaction  and drug-target interaction prediction.

\begin{acks}
This work was supported in part by the National Natural Science Foundation of China under Contract 61836011 and in part by the Youth Innovation Promotion Association CAS under Grant 2018497. 
\end{acks}

\bibliographystyle{ACM-Reference-Format}
\bibliography{mybib}

\appendix

\section{Details about the network}
\label{app:network_details_app}

As introduced in Section \ref{sec:net_arch}, the model is a stack of $L$ identical blocks, where each block takes the output of the previous one as input. The $(l-1)$-th block will output the representations of atom $v_i$ (denoted as $x^{(l-1)}_i\in\mathbb{R}^d$), the representations of bond $e_{ij}$ (denoted as $x_{ij}^{(l-1)}\in\mathbb{R}^d$), a predicted conformation $\hat{R}^{(l-1)}$, and a global representation of the molecule $u^{(l-1)}$.  Let $\mathcal{N}(i)$ denote the neighbors of atom $i$, i.e., $\mathcal{N}(i)=\{j\mid e_{ij}\in E\}$.The workflow of the $l$-th block is shown as follows:

\noindent(1) {\em Incorporate geometric information}:
\begin{equation}
\begin{aligned}
& \bar{x}^{(l)}_i = x^{(l-1)}_i + \texttt{FF}(\hat{R}^{(l-1)}_i),\\
& \bar{x}^{(l)}_{ij} = x^{(l-1)}_{ij} + \texttt{FF}(\Vert \hat{R}^{(l-1)}_i- \hat{R}^{(l-1)}_j \Vert).
\end{aligned}
\end{equation}
\noindent(2) {\em Update bond representations}: For each $e_{ij}\in E$, 
\begin{equation}
x^{(l)}_{ij} = x^{(l-1)}_{ij} + \texttt{FF}(\bar{x}^{(l-1)}_i, \bar{x}^{(l-1)}_j, \bar{x}^{(l)}_{ij},u^{(l-1)}).
\end{equation}
\noindent(3) {\em Update atom representations}: For any $i\in V$,
\begin{equation}
\begin{aligned}
& \tilde{x}_i^{(l)}=\sum_{j\in\mathcal{N}(i)}\alpha_jW_v[x^{(l)}_{ij};\bar{x}_j^{(l)}];\\
& \alpha_j\propto\exp(\bm{a}^\top \xi(W_q\bar{x}^{(l-1)}_i+W_k[\bar{x}^{(l-1)}_j; \bar{x}^{(l)}_{ij}])  );\\
& x^{(l)}_{i}= x_i^{(l-1)} + \texttt{MLP}\Big(x^{(l-1)}_i, \tilde{x}_i^{(l)}, u^{(l-1)}\Big).
\end{aligned}
\label{eq:update_atom_feature}
\end{equation}
In Eqn.\eqref{eq:update_atom_feature},  $\bm{a}$, $W_q$, $W_v$ and $W_k$ are the parameters to be learned, $[a;b]$ is the concatenation of two vectors $a$ and $b$, and $\xi$ is the leaky ReLU activation.

\noindent(4) {\em Update global representations}: The global representations are updated by aggregating the new bond representations and atom representations, i.e.,
\begin{equation}
u^{(l)} = u^{(l-1)} + \texttt{FF}\Big(\frac{1}{|V|}\sum_{i=1}^{|V|}x^{(l)}_i, \frac{1}{|E|}\sum_{i,j}x^{(l)}_{ij},u^{(l-1)}\Big).
\label{eq:update_global_feature}
\end{equation}
After obtaining the updated representations, the $l$-th block predicts a new 3D conformation. For any $i\in\{1,2,\cdots,|V|\}$,
\begin{equation}
\begin{aligned}
& \Delta^{(l)}_i=\texttt{FF}(x^{(l)}_i),\; \mu^{(l)}=\frac{1}{|V|}\sum_{j=1}^{|V|}\Delta^{(l)}_j,\\
&\hat{R}^{(l)}_i=\hat{R}^{(l)}_i + (\Delta^{(l)}_i-\mu^{(l)}).
\label{eq:update_conformer}
\end{aligned}
\end{equation}

\begin{table}[!htbp]
\centering
\small
\begin{tabular}{lcc}
\toprule
Hyper parameters & MoleculeNet & Toxicity \\
\midrule
Learning Rate&\{2e-4, 5e-4, 8e-4, 1e-3\}& \{5e-4, 8e-4, 1e-3, 2e-3\} \\
Dropout  &\{0.3, 0.4, 0.5\}&\{0.2, 0.3, 0.4, 0.5\}\\
Batch Size &\{16, 32, 64, 128\}&\{16, 32, 64\}\\
Max Epochs&\{10, 20, 50\} &\{10, 20, 50, 100\}\\
Weight Decay &\{ 0.01, 0.1\}& \{0.01, 0.1\}\\
\bottomrule
\end{tabular}
\caption{Finetuning hyperparameters.}
\label{tab:fineutning_hyperparams}
\end{table}

\section{Finetuning Hyperparameters}\label{sec:morehyperparams}
We summarize the finetuning hyperparameters in Table \ref{tab:fineutning_hyperparams}. The hyperparameters for the six tasks from MoleculeNet are in the ``MoleculeNet'' column, while those for the four toxicity prediction tasks are in the ``Toxicity'' column. 

The details of the dataset are in Table \ref{tab:dataset_description}.

\section{Pre-training with different loss function}\label{sec:diff_loss}

\begin{table}[!htpb]
\centering
\begin{tabular}{lccccccc}
\toprule
Dataset& BBBP  & Tox21 &ClinTox &  \\
\midrule
$\ell_{\rm atom}+\ell_{\rm coord}$& $76.0\pm0.5$&$75.5\pm0.4$&$91.3\pm2.5$\\
$\ell_\textrm{2D$\to$3D}$&$75.8\pm0.8$&$74.7\pm0.6$&$89.4\pm3.4$\\
$\ell_\textrm{3D$\to$2D}$&$75.5\pm0.7$&$75.0\pm0.4$& $93.4\pm1.6$\\
Ours &$77.3\pm0.4$& $75.9\pm0.3$&$95.0\pm1.1$\\
\midrule
Dataset& HIV  & BACE & SIDER  \\
\midrule
$\ell_{\rm atom}+\ell_{\rm coord}$& $81.2\pm1.4$&$85.1\pm0.5$&$65.8\pm1.3$ \\
$\ell_\textrm{2D$\to$3D}$& $81.0\pm1.3$&$85.3\pm1.3$&$65.2\pm0.8$  \\
$\ell_\textrm{3D$\to$2D}$ &  $80.9\pm1.2$& $85.4\pm1.5$&$65.1\pm1.5$ \\
Ours  &$82.2\pm1.0$&$86.8\pm0.6$&$67.4\pm0.5$ & \\
\bottomrule
\end{tabular}
\caption{Ablation study of different training objective functions on MoleculeNet.}
\label{tab:moleculenet_ablation}
\end{table}

As introduced in Section \ref{sec:ablation}, we conduct ablation study of pre-training with different objective functions, and finetuning on MoleculeNet. The results are summarized in Table~\ref{tab:moleculenet_ablation}. We can see that independently using any objective is not as good as combining them together.

\begin{table*}[!b]
\small
\centering
\caption{Details about the datasets for molecular property prediction.}
\begin{tabular}{lcclc}
\toprule
Dataset & 3D & \#Instance & Description & Evaluation \\
\midrule 
BBBP & N & 2053 & Prediction of the barrier permeability of the small molecules. & ROC-AUC \\
Tox21 & N & 8014 &  Qualitative
toxicity measurements on 12 different targets. & ROC-AUC \\
ClinTox & N & 1491 & The prediction of clinical trial toxicity and FDA approval status. & ROC-AUC \\
HIV & N & 41913 & Prediction of the ability to inhibit HIV replication. & ROC-AUC \\
BACE & N & 1522 &  Prediction of binding results for a set of inhibitors of human $\beta$-secretase 1. & ROC-AUC \\
SIDER & N & 1427 & The side effect of drugs. & ROC-AUC \\
PCBA & N & 437929 & Prediction of biological activities obtained by HTS. & Average Precision (AP) \\
LD50 & Y & 7413 & The amount of chemicals that can kill half of the rats when orally ingested. & $R^2$/RMSE/MAE \\
LGC50 & Y & 1792 & 50\% growth inhibitory concentration of Tetrahymena pyriformis organism after 40h. & $R^2$/RMSE/MAE  \\
LC50 & Y & 823 & The concentration of test chemicals in water that causes 50\% of fathead minnows to die after 96h. & $R^2$/RMSE/MAE \\
LC50DM & Y & 353 & The concentration of test chemicals in water that causes 50\% Daphnia Magna to die after 48h. & $R^2$/RMSE/MAE \\
\bottomrule
\end{tabular}
\label{tab:dataset_description}
\end{table*}

\end{document}